\def\year{2020}\relax
\documentclass[letterpaper]{article} % DO NOT CHANGE THIS
\usepackage{aaai20}  % DO NOT CHANGE THIS
\usepackage{times}  % DO NOT CHANGE THIS
\usepackage{helvet} % DO NOT CHANGE THIS
\usepackage{courier}  % DO NOT CHANGE THIS
\usepackage[hyphens]{url}  % DO NOT CHANGE THIS
\usepackage{graphicx} % DO NOT CHANGE THIS
\usepackage{graphicx}  % DO NOT CHANGE THIS
\usepackage{soul}
\usepackage[hidelinks]{hyperref}
\usepackage[utf8]{inputenc}
\usepackage[small]{caption}
\usepackage{amsmath}
\usepackage{amssymb}
\usepackage{booktabs}
\usepackage{algorithm}
\usepackage{algorithmic}
\usepackage{multirow}
\usepackage{color}
\title{Lexical Simplification with Pretrained Encoders }
\author{  \Large \textbf{Jipeng Qiang$^1$, Yun Li$^1$, Yi Zhu$^1$, Yunhao Yuan$^1$, Xindong Wu$^{2,3}$}\\ % All authors must be in the same font size and format. Use \Large and \textbf to achieve this result when breaking a line
\textsuperscript{\rm 1}Department of Computer Science, Yangzhou University, Jiangsu, China\\
\textsuperscript{\rm 2}Key Laboratory of Knowledge Engineering with Big Data (Hefei University of Technology), Ministry of Education, Anhui, China\\
\textsuperscript{\rm 3} Mininglamp Academy of Sciences, Minininglamp Technology, Beijing, China\\
\{jpqiang,liyun,zhuyi,yhyuan\}@yzu.edu.cn, xwu@hfut.edu.cn % email address must be in roman text type, not monospace or sans serif
}

\begin{document}

\maketitle

\begin{abstract}
Lexical simplification (LS) aims to replace complex words in a given sentence with their simpler alternatives of equivalent meaning. Recently unsupervised lexical simplification approaches only rely on the complex word itself regardless of the given sentence to generate candidate substitutions, which will inevitably produce a large number of spurious candidates. We present a simple LS approach that makes use of the Bidirectional Encoder Representations from Transformers (BERT) which can consider both the given sentence and the complex word during generating candidate substitutions for the complex word. Specifically, we mask the complex word of the original sentence for feeding into the BERT to predict the masked token. The predicted results will be used as candidate substitutions. Despite being entirely unsupervised, experimental results show that our approach obtains obvious improvement compared with these baselines leveraging linguistic databases and parallel corpus, outperforming the state-of-the-art by more than 12 Accuracy points on three well-known benchmarks.

\end{abstract}

\section{Introduction} 

Lexical Simplification (LS) aims at replacing complex words with simpler alternatives, which can help various groups of people, including children \cite{De_belder}, non-native speakers \cite{paetzold2016unsupervised}, people with cognitive disabilities \cite{feng2009automatic,saggion2017automatic}, to understand text better. The popular LS systems still predominantly use a set of rules for substituting complex words with their frequent synonyms from carefully handcrafted databases (e.g., WordNet) or automatically induced from comparable corpora \cite{devlin1998the,De_belder}. Linguistic databases like WordNet are used to produce simple synonyms of a complex word. \cite{Lesk:1986:ASD:318723.318728,sinha2012unt,leroy2013user}. Parallel corpora like Wikipedia-Simple Wikipedia corpus were also used to extract complex-to-simple word correspondences \cite{biran2011putting,yatskar2010sake,horn2014learning}. However, linguistic resources are scarce or expensive to produce, such as WordNet and Simple Wikipedia, and it is impossible to give all possible simplification rules from them. 

\begin{figure}
  \centering
  \includegraphics[width=70mm]{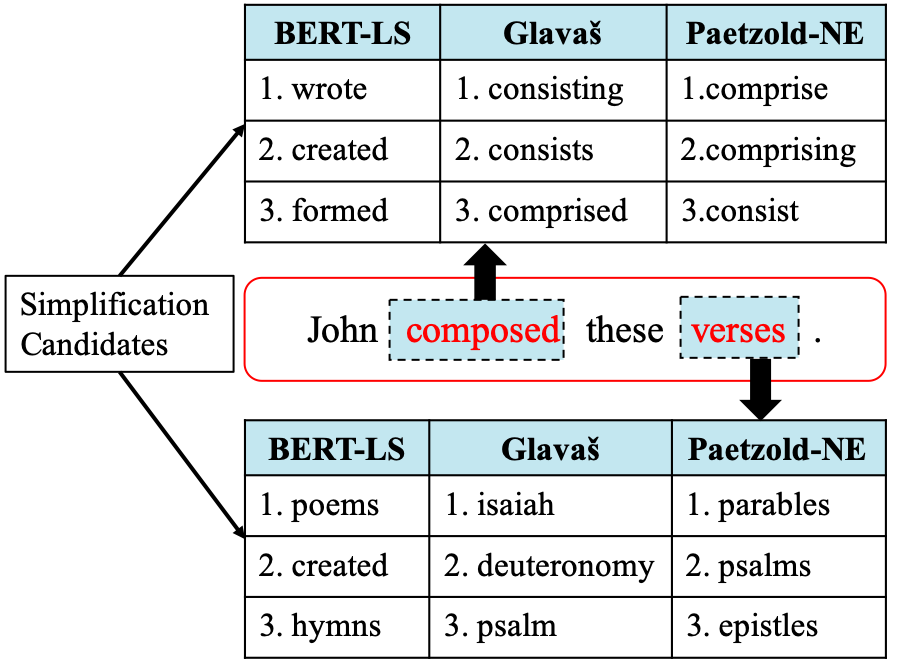}
  \caption{Comparison of simplification candidates of complex words. Given one sentence "John composed these verses." and complex words 'composed' and 'verses', the top three simplification candidates for each complex word are generated by our method BERT-LS and the state-of-the-art two baselines based word embeddings (Glava{\v{s}}\protect\cite{glavavs2015simplifying} and Paetzold-NE \protect\cite{paetzold2017lexical}).}
  \label{Fig1}
\end{figure}

For avoiding the need for resources such as databases or parallel corpora, recent work utilizes word embedding models to extract simplification candidates for complex words \cite{glavavs2015simplifying,paetzold2016unsupervised,paetzold2017lexical}. Given a complex word, they extracted from the word embedding model the simplification candidates whose vectors are closer in terms of cosine similarity with the complex word. This strategy achieves better results compared with rule-based LS systems. However, the above methods generated simplification candidates only considering the complex word regardless of the context of the complex word, which will inevitably produce a large number of spurious candidates that can confuse the systems employed in the subsequent steps.

Therefore, we present an intuitive and innovative idea completely different from existing LS systems in this paper. We exploit recent advances in the pre-trained transformer language model BERT \cite{devlin2018bert} to find suitable simplifications for complex words. The masked language model (MLM) used in BERT randomly masks some percentage of the input tokens, and predicts the masked word based on its context. If masking the complex word in a sentence, the idea in MLM is in accordance with generating the candidates of the complex word in LS. Therefore, we introduce a novel LS approach BERT-LS that uses MLM of BERT for simplification candidate generation. More specifically, we mask the complex word $w$ of the original sentence $S$ as a new sentence $S'$, and we concatenate the original sequence $S$ and $S'$ for feeding into the BERT to obtain the probability distribution of the vocabulary corresponding to the masked word. The advantage of our method is that it generates simplification candidates by considering the whole sentence, not just the complex word.

Here, we give an example shown in Figure 1 to illustrate the advantage of our method BERT-LS. For complex words 'composed' and 'verses' in the sentence "John composed these verses.", the top three substitution candidates of the two complex words generated by the LS systems based on word embeddings \cite{glavavs2015simplifying,paetzold2017lexical} are only related with the complex words itself without without paying attention to the original sentence. The top three substitution candidates generated by BERT-LS are not only related with the complex words, but also can fit for the original sentence very well. Then, by considering the frequency or order of each candidate, we can easily choose 'wrote' as the replacement of 'composed and 'poems' as the replacement of 'verses'. In this case, the simplification sentence 'John wrote these poems.' is more easily understood than the original sentence. 

The contributions of our paper are as follows:

(1) BERT-LS is a novel BERT-based method for LS, which can take full advantages of BERT to generate and rank substitution candidates. Compared with existing methods, BERT-LS is easier to hold cohesion and coherence of a sentence, since BERT-LS considers the whole sentence not the complex word itself during generating candidates.

(2) BERT-LS is a simple, effective and unsupervised LS method. 1)Simple: many steps used in existing LS systems have been eliminated from our method, e.g., morphological transformation and substitution selection. 2) Effective: it obtains new state-of-the-art results on three benchmarks. 3) Unsupervised: our method cannot rely on any parallel corpus or linguistic databases.

(3) To our best knowledge, this is the first attempt to apply Pre-Trained Transformer Language Models on lexical simplification tasks. The code to reproduce our results is available at https://github.com/anonymous.

\section{Related Work}

Lexical simplification (LS) contains identifying complex words and finding the best candidate substitution for these complex words \cite{shardlow2014survey,paetzold2017survey}.
The best substitution needs to be more simplistic while preserving the sentence grammatically and keeping its meaning as much as possible, which is a very challenging task. The popular lexical simplification (LS) approaches are rule-based, which each rule contain a complex word and its simple synonyms \cite{Lesk:1986:ASD:318723.318728,pavlick2016simple,maddela-xu-2018-word}. In order to construct rules, rule-based systems usually identified synonyms from WordNet for a predefined set of complex words, and selected the "simplest" from these synonyms based on the frequency of word \cite{devlin1998the,De_belder} or length of word \cite{bautista2011empirical}. However, rule-based systems need a lot of human involvement to manually define these rules, and it is impossible to give all possible simplification rules.

As complex and simplified parallel corpora are available, especially, the 'ordinary' English Wikipedia (EW) in combination with the 'simple' English Wikipedia (SEW), the paradigm shift of LS systems is from knowledge-based to data-driven simplification \cite{biran2011putting,yatskar2010sake,horn2014learning}. Yatskar et al. (2010) identified lexical simplifications from the edit history of SEW. They utilized a probabilistic method to recognize simplification edits distinguishing from other types of content changes. Biran et al. (2011) considered every pair of the distinct word in the EW and SEW to be a possible simplification pair, and filtered part of them based on morphological variants and WordNet. Horn et al. (2014) also generated the candidate rules from the EW and SEW, and adopted a context-aware binary classifier to decide whether a candidate rule should be adopted or not in a certain context. The main limitation of the type of methods relies heavily on simplified corpora. 

In order to entirely avoid the requirement of lexical resources or parallel corpora, LS systems based on word embeddings were proposed \cite{glavavs2015simplifying}. They extracted the top 10 words as candidate substitutions whose vectors are closer in terms of cosine similarity with the complex word. Instead of a traditional word embedding model, Paetzold and Specia (2016) adopted context-aware word embeddings trained on a large dataset where each word is annotated with the POS tag. Afterward, they further extracted candidates for complex words by combining word embeddings with WordNet and parallel corpora \cite{paetzold2017lexical}. 

After examining existing LS methods ranging from rules-based to embedding-based, the major challenge is that they generated simplification candidates for the complex word regardless of the context of the complex word, which will inevitably produce a large number of spurious candidates that can confuse the systems employed in the subsequent steps. 

In this paper, we will first present a BERT-based LS approach that requires only a sufficiently large corpus of regular text without any manual efforts. Pre-training language models \cite{devlin2018bert,lee2019biobert,lample2019cross} have attracted wide attention and has shown to be effective for improving many downstream natural language processing tasks. Our method exploits recent advances in BERT to generate suitable simplifications for complex words. Our method generates the candidates of the complex word by considering the whole sentence that is easier to hold cohesion and coherence of a sentence. In this case, many steps used in existing steps have been eliminated from our method, e.g., morphological transformation and substitution selection. Due to its fundamental nature, our approach can be applied to many languages.

\section{Unsupervised Lexical Simplification}

In this section, we briefly summarize the BERT model, and then describe how we extend it to do lexical simplification.

\subsection{The BERT model}

BERT \cite{devlin2018bert} is trained on a masked language modeling (MLM) objective, which is combination of a Markov Random Field language model on a set of discrete tokens with pseudo log-likelihood \cite{wang2019bert}. Unlike a traditional language modeling objective of predicting the next word in a sequence given the history, masked language modeling predicts a word given its left and right context. Let $S={w_1,...,w_l,...,w_L}$ be a set of discrete tokens, $w_l\in V$, where $V=\{v_1,...,v_M\}$ is the vocabulary. A joint probability distribution of a given sequence $X$ can be calculated by:

\[
\begin{split}
p(S|\theta)=\frac{1}{Z(\theta)}\prod_{l=1}^{L}\phi_l(w|\theta)\propto exp(\sum_{l=1}^{L}\log\phi_l(w|\theta))
\end{split}
\]
where $\phi_l(w)$ is the $l$'th potential function with parameters $\theta$, and $Z$ is the partition function.

The log potential functions for the $l$'th token are defined by:

\[
\begin{split}
\log \phi_l(w|\theta)=w_l^Tf_{\theta}(S_{\backslash l})
\end{split}
\]
where $w_l$ is one-hot vector for the $l$'the token, and $S_{\backslash l} =(w_1,...,w_{l-1},$[MASK],$w_{l+1},...,w_L)$

The function $f(S_{\backslash l})\in \mathbb{R}^M$ is a multi-layer bidirectional transformer model \cite{vaswani2017attention}. See \cite{devlin2018bert} for details. The model is trained to approximately maximize the pseudo log-likelihood

\[
\begin{split}
L(\theta) = E_{S\sim D} \sum_{l=1}^{L}\log p(w_l|S_{\backslash l};\theta)
\end{split}
\]
where $D$ is a set of training examples. In practice, we can stochastically optimize the logloss (computed from the softmax predicted by the $f$ function) by sampling tokens as well as training sentences.

Besides, BERT also uses a “next sentence prediction” task that jointly pre-trains text-pair representations. BERT accomplishes this by prepending every sentence with a special classification token, [CLS], and by combining sentences with a special separator token, [SEP]. The final hidden state corresponding to the [CLS] token is used as the total sequence representation from which we predict a label for classification tasks, or which may otherwise be overlooked.

\subsection{Simplification Candidate generation}

Simplification candidate generation is used to produce candidate substitutions for each complex word. Due to the fundamental nature of masked language modeling (MLM), we mask the complex word $w$ of the sentence $S$ and get the probability distribution of the vocabulary $p(\cdot|S\backslash \{w\})$ corresponding to the masked word $w$ using MLM. Therefore, we can try to use MLM for simplification candidate generation.

For the complex word $w$ in a sentence $S$, we mask the word $w$ of $S$ as a new sequence $S'$. If we directly feed $S'$ into MLM, the probability of the vocabulary $p(\cdot|S'\backslash \{t_i\})$ corresponding to the complex word $w$ only considers the context regardless of the influence of the complex word $w$.

\begin{figure*}
  \centering
  \includegraphics[width=170mm]{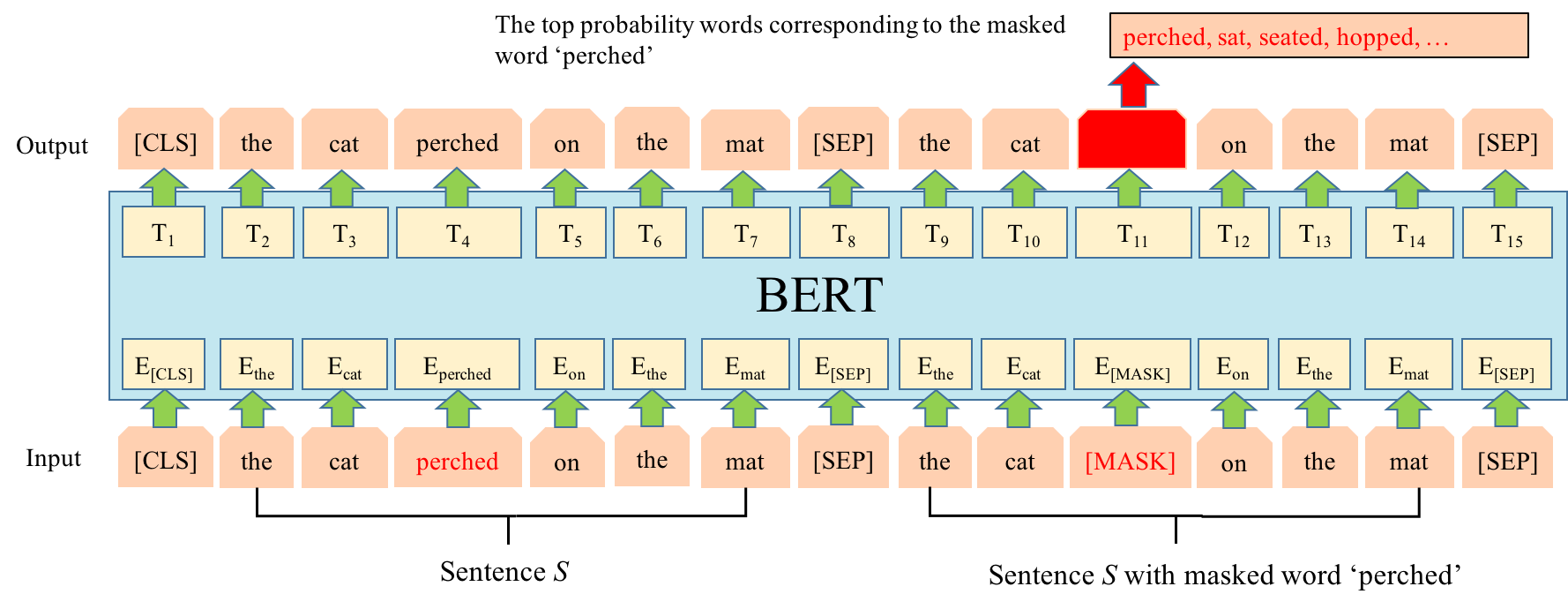},
  \caption{Substitution generation of BERT-LS for the target complex word prediction, or $cloze$ task. The input text is "the cat perched on the mat" with complex word "perched". [CLS] and [SEP] are two special symbols in BERT, where [CLS] the token for classification (will not be used in this paper) and [SEP] is a special separator token. }
  \label{MLM}
\end{figure*} 
 
Since BERT is adept at dealing with sentence pairs due to the "next sentence prediction" adopted by BERT. We concatenate the original sequence $S$ and $S'$ as a sentence pair, and feed the sentence pair ${S,S'}$ into the BERT to obtain the probability distribution of the vocabulary $p(\cdot|S,S'\backslash \{w\})$ corresponding to the mask word. In this way, the higher probability words in $p(\cdot|S,S'\backslash \{w\})$ corresponding to the mask word not only consider the complex word itself, but also can fit the context of the complex word. Finally, we select as simplification candidates the top 10 words from $p(\cdot|S,S'\backslash \{w\})$, excluding the morphological derivations of $w$. In all experiments, we use BERT-Large, Uncased (Whole Word Masking) pre-trained on BooksCorpus and English Wikipedia \footnote{https://github.com/google-research/bert}.

Suppose that there is a sentence "the cat perched on the mat" and the complex word "perched", we can get the top three simplification candidate words "sat, seated, hopped". See Figure \ref{MLM} for an illustration. We can see the three candidates not only have a strong correlation with the complex word, but also hold the cohesion and coherence properties of the sentence. If we adopt the existing state-of-the-art method based on word embeddings proposed by \cite{glavavs2015simplifying}, the top three substitution words are "atop, overlooking, precariously". Very obviously, our method generates better candidates.

\subsection{Substitution Ranking}

The substitution ranking of the lexical simplification pipeline is to decide which of the candidate substitutions that fit the context of a complex word is the simplest \cite{paetzold2017survey}. We rank the candidate substitutions based on the following features. Each of the features captures one aspect of the suitability of the candidate word to replace the complex word. 

\textbf{BERT prediction.} On this step of simplification candidate generation, we obtain the probability distribution of the vocabulary corresponding to the mask word $p(\cdot|S,S'\backslash \{w\})$. The higher the probability, the more relevant the candidate for the original sentence. The candidate substitutions can be ranked according to their probability.

\textbf{Language model features.} A simplification candidate should fit into the sequence of words preceding and following the original word. Different n-gram language model, we cannot compute the probability of a sentence or sequence of words using MLM. Let $W=w_{-m},...,w_{-1},w,w_1,...,w_m$ be the context of the original word $w$. We adopt a new strategy to compute the likelihood of $W$. We first replace the original word $w$ into the simplification candidate. We then mask one word of $W$ from front to back, and feed into MLM to compute the cross-entropy loss. Finally, we rank all simplification candidates based on the average loss of $W$. The lower the loss, the simplification candidate is a good substitute for the original word. We use as context a symmetric window of size five around the complex word.

\textbf{Semantic similarity.} The feature is calculated as the cosine between the fastText \footnote{https://dl.fbaipublicfiles.com/fasttext/vectors-english/crawl-300d-2M-subword.zip} vector of the original word and the fastText vector of the candidate substitutions. The higher the similarity, the more similar the two words.

\textbf{Frequency feature.} Frequency-based candidate ranking strategies are one of the most popular choices by lexical simplification and can be quite effective. In general, the more frequency a word is used, the most familiar it is to readers. We use word frequency estimates from the Wikipedia \footnote{https://dumps.wikimedia.org/enwiki/} and the Children’s Book Test (CBT) \footnote{The dataset can be downloaded from http://fb.ai/babi/}. As the size of Wikipedia is larger than CBT, we only choose the top 12 million texts of Wikipedia for matching the size.

\subsection{Simplification Algorithm}

The overall simplification algorithm BERT-LS is shown in Algorithm 1. In this paper, we are not focused on identifying complex words \cite{paetzold2017survey}, which is a separate task. For each complex word, we first get simplification candidates using BERT after preprocessing the original sequence (lines 1-4). Afterward, BERT-LS computes various rankings for each of the simplification candidates using each of the features, and then scores each candidate by averaging all its rankings (lines 5-13). We choose the top candidate with the highest average rank over all features as the simplification replacement (line 15). 

\begin{algorithm}[tb]
\caption{ Simplify(sentence $S$, Complex word $w$)}
\label{alg:algorithm}
\begin{algorithmic}[1] %[1] enables line numbers
\STATE Replace word $w$ of $S$ into [MASK] as $S'$
\STATE Concatenate $S$ and $S'$ using [CLS] and [SEP]
\STATE $p(\cdot|S,S'\backslash \{w\}) \leftarrow BERT(S,S')$
\STATE $scs \leftarrow top\_probability(p(\cdot|S,S'\backslash \{w\}))$
\STATE $all\_ranks$ $\leftarrow$ $\varnothing$
\FOR{ each feature $f$ }
\STATE $scores$ $\leftarrow \varnothing$
\FOR{ each $sc\in scs$ }
\STATE $scores$ $\leftarrow$ $scores$ $\cup$ $f(sc)$
\ENDFOR
\STATE $rank$ $\leftarrow$ $rank\_numbers(scores)$
\STATE $all\_ranks$ $\leftarrow$ $all\_ranks$ $\cup$ $rank$
\ENDFOR
\STATE $avg\_rank$ $\leftarrow$ $average(all\_ranks)$
\STATE $best$ $\leftarrow$ argmax$_{sc}(avg\_rank)$
\RETURN $best$
\end{algorithmic}
\end{algorithm}

\section{Experiments}

We design experiments to answer the following questions:

\textbf{Q1. The effectiveness of simplification candidates:} Does the simplification candidate generation of BERT-LS outperforms the substitution generation of the state-of-the-art competitors?

\textbf{Q2. The effectiveness of the LS system: } Do the FULL Pipeline of BERT-LS outperforms the full pipeline of the state-of-the-art competitors?

\subsection{Experiment Setup}

\begin{table*}
\centering
\begin{tabular}{l|ccc|ccc|ccc|}
\hline
& \multicolumn{3}{|c|}{LexMTurk}  & \multicolumn{3}{|c|}{BenchLS}  & \multicolumn{3}{|c|}{NNSeval} \\

 &   Precision & Recall & F1 & Precision & Recall & F1 & Precision & Recall & F1\\
\hline
Yamamoto & 0.056 &  0.079 & 0.065  & 0.032 & 0.087 & 0.047 & 0.026 & 0.061 &  0.037 \\
Devlin & 0.164 &  0.092 & 0.118 & 0.133 & 0.153 & 0.143 & 0.092 & 0.093 & 0.092  \\
Biran &  0.153 & 0.098  & 0.119  & 0.130 & 0.144 & 0.136 & 0.084 & 0.079 & 0.081  \\
Horn &  0.153 & 0.134  & 0.143 & 0.235 & 0.131 & 0.168 & 0.134 & 0.088 & 0.106  \\
Glava{\v{s}} &  0.151 &  0.122 &  0.135 & 0.142 & 0.191 & 0.163 & 0.105 & 0.141 & 0.121 \\ 
Paetzold-CA &  0.177 & 0.140  & 0.156  & 0.180 & 0.252 & 0.210 & 0.118 & 0.161 & 0.136 \\
Paetzold-NE &  \textbf{0.310} & 0.142  &  0.195 & \textbf{0.270} & 0.209 & 0.236 & 0.186 & 0.136 & 0.157 \\
\hline
BERT-LS &  0.296 & \textbf{0.230} & \textbf{0.259} & 0.236 & \textbf{0.320} & \textbf{0.272} &  \textbf{0.190} & \textbf{0.254} & \textbf{0.218} \\
\hline
\end{tabular}
\caption{Evaluation results of simplification candidate generation on three datasets. }
\label{SGResults}
\end{table*}

\textbf{Dataset}. We use three widely used lexical simplification datasets to do experiments. 

(1) LexMTurk\footnote{http://www.cs.pomona.edu/~dkauchak/simplification/lex.mturk.14} \cite{horn2014learning}. It is composed of 500 instances for English. Each instance contains one sentence from Wikipedia, a target complex word and 50 substitutions annotated by 50 Amazon Mechanical "turkers". Since each complex word of each instance was annotated by 50 turkers, this dataset owns a pretty good coverage of gold simplifications.

(2) BenchLS \footnote{http://ghpaetzold.github.io/data/BenchLS.zip} \cite{paetzold2016unsupervised}. It is composed of 929 instances for English, which is from LexMTurk and LSeval \cite{De_belder}. The LSeval contains 429 instances, in which each complex word was annotated by 46 turkers and 9 Ph.D. students. Because BenchLS contains two datasets, it provides the largest range of distinct complex words amongst the LS dataset of English.

(3) NNSeval \footnote{http://ghpaetzold.github.io/data/NNSeval.zip} \cite{paetzold2017survey}. It is composed of 239 instances for English, which is a filtered version of BenchLS. These instances of BenchLS are dropped: 1) the target complex word in the instance was not regarded as the complex word by a non-native English speaker, and 2) any candidate in the instance was regarded as a complex word by a non-native speaker. 

Notice that, because these datasets already offer target words regarded complex by human annotators, we do not address complex word identification task in our evaluations.

\textbf{Comparison Systems}. We choose the following eight baselines to evaluation: Devlin \cite{devlin1998the}, Biran \cite{biran2011putting}, Yamamoto \cite{kajiwara2013selecting}, Horn \cite{horn2014learning}, Glava{\v{s}} \cite{glavavs2015simplifying}, SimplePPDB \cite{pavlick2016simple}, Paetzold-CA \cite{paetzold2016unsupervised}, and Paetzold-NE \cite{paetzold2017lexical}. The substitution generation strategies of these baselines generate candidates from WordNet, EW and SEW parallel corpus, Merriam dictionary, EW and SEW parallel corpus, word embeddings, context-aware word embeddings, combining the Newsela parallel corpus and context-aware word embeddings.

\subsection{Simplification Candidate Generation}

The following three widely used metrics are used for evaluation \cite{paetzold2015lexenstein,paetzold2016unsupervised}. 

\textbf{Precision}: The proportion of generated candidates that are in the gold standard. 

\textbf{Recall}: The proportion of gold-standard substitutions that are included in the generated substitutions. 

\textbf{F1}: The harmonic mean between Precision and Recall.

The results are shown in Table \ref{SGResults}. The results of the baselines on LexMTurk are from \cite{paetzold2017survey} and the results on BenchLS and NNSeval are from \cite{paetzold2017lexical}. As can be seen, despite being entirely unsupervised, our model BERT-LS obtains F1 scores on three datasets, largely outperforming the previous best baselines. The baseline Paetzold-NE by combining the Newsela parallel corpus and context-aware word embeddings obtains better results on Precision metric, which demonstrates that substitution generation tends to benefit from the combination of different resources. But, for under-resourced languages, our method is still likely to create competitive candidate generators. The results clearly show that BERT-LS provides a good balance precision and recall using only BERT over raw text.

\begin{table*}
\centering
\begin{tabular}{|l|l|}

 \hline
Sentence & it originally \textcolor{red}{aired} on the Fox network in the United States on October 10, 1991. \\
Labels & showed, played, shown, appeared, ran, televised, broadcasted, premiered, opened\\
Glava{\v{s}} & \textbf{broadcasted}, broadcast, re-aired, taped, unbroadcast, al-sumariya \\
Paetzold-NE & broadcasting, taped, \textbf{broadcasted}, re-broadcast, telecast, re-air\\
BERT-LS & \textbf{appeared}, \textbf{ran}, \textbf{premiered}, \textbf{opened}, screened, broadcast  \\
 \hline
Sentence & a good measure of notability is whether someone has been featured in \textcolor{red}{multiple}, independent sources.\\
Labels & many, several, different, numerous, trotting, some \\
Glava{\v{s}} & simultaneous, discrete, simultaneously, predefined, types, overlapping \\
Paetzold-NE & discrete, simultaneous, overlapping, pre-defined, various, other\\
BERT-LS & \textbf{several}, \textbf{numerous}, two, \textbf{many}, \textbf{different}, single \\
\hline
Sentence & he normally appears in a running gag, where he usually suffers \textcolor{red}{unfortunate}, nearly always fatal, events.\\
Labels & bad, unlucky, sad, terrible, unhappy, hapless, tragic, unlikely, damaging, disasterous \\
Glava{\v{s}} & inexplicable, regrettable, distressing, lamentable, galling, obvious\\
Paetzold-NE & lamentable, galling, obvious, dreadful, inexplicable, \textbf{terrible}\\
BERT-LS & \textbf{terrible}, unacceptable, exceptional, unpleasant, \textbf{tragic}, unexpected\\
\hline
Sentence & Tolstoy was born in Yasnaya Polyana, the family \textcolor{red}{estate} in the Tula region of Russia. \\
Labels & home, property, house, residence, ands, land, domain, grounds \\
Glava{\v{s}} & \textbf{property}, realty, freehold, klondikes, real-estate, estate- \\
Paetzold-NE & condominium, leasehold, realty, xf.com, financier-turned-felon, mirvac\\
BERT-LS & \textbf{property}, \textbf{house}, seat, plantation, farm, station\\
\hline
Sentence & Gygax is \textcolor{red}{generally} acknowledged as the father of role-playing games. \\
Labels & usually, normally, mainly, often, mostly, widely, commonly, traditionally, roughly\\
Glava{\v{s}} & however, are, therefore, \textbf{usually}, furthermore, conversely\\
Paetzold-NE & therefore, furthermore, uniformly, moreover, as, reasonably\\
BERT-LS & \textbf{widely}, \textbf{usually}, \textbf{commonly}, \textbf{often}, typically, largely \\
\hline
Sentence & he is notable as a \textcolor{red}{performer} of Australian traditional folk songs in an authentic style.\\
Labels & singer, entertainer, musician, artist, player, actor, worker\\
Glava{\v{s}} & soloist, pianist, vocalist, flautist, \textbf{musician}, songwriter \\
Paetzold-NE & showman, soloist, pianist, dancer, vocalist, \textbf{musician}\\
BERT-LS & \textbf{singer}, practitioner, vocalist, soloist, presenter, \textbf{player} \\
\hline
Sentence & Aung San Suu Kyi returned to Burma in 1988 to take care of her \textcolor{red}{ailing} mother.\\
Labels & sick, ill, sickly, dying, suffering, distressed, sickening \\
Glava{\v{s}} & troubled, foundering, beleaguered, cash-starved, cash-strapped, debt-ridden\\
Paetzold-NE & embattled, cash-strapped, beleaguered, foundering, struggling, debt-laden\\
BERT-LS & \textbf{dying}, \textbf{sick}, \textbf{ill}, elderly, injured, ai \\
\hline
Sentence & the Convent has been the official \textcolor{red}{residence} of the Governor of Gibraltar since 1728.    \\
Labels &home, house, dwelling, abode, owner, people  \\
Glava{\v{s}} & dormitory, domicile, haseldorf, accommodation, non-student, mirrielees\\
Paetzold-NE & 15-bathroom, dormitory, ashenhurst, student-housing, storthes, domicile\\
BERT-LS & \textbf{home}, seat, \textbf{house}, mansion, housing, haven \\
\hline
Sentence& Kowal suggested the name and the IAU \textcolor{red}{endorsed} it in 1975.    \\
Labels& supported, approved, accepted, backed, okayed, authorized, favored, passed, adopted, ratified \\
Glava{\v{s}} & \textbf{adopted}, \textbf{backed}, rejected, unanimously, drafted, \textbf{supported} \\
Paetzold-NE &drafted, advocated, lobbied, rejected, championed, \textbf{supported} \\
BERT & \textbf{approved}, \textbf{adopted}, sanctioned, \textbf{supported}, \textbf{ratified}, \textbf{backed}\\
\hline
Sentence & it overwinters in \textcolor{red}{conifer} groves . \\
Labels& tree, pine, evergreen, fir, wood, cedar, grassy, nice\\
Glava{\v{s}} & redcedar, groundlayer, multistemmed, coniferous, needle-leaf, deciduous \\
Paetzold-NE & oak-hickory, castanopsis, pseudotsuga, douglas-fir, menziesii, alnus\\
BERT-LS & \textbf{pine}, cypress, \textbf{fir}, redwood, \textbf{cedar}, forested\\
\hline
\end{tabular}
\caption{Examples of various sentences from LexMTurk dataset. The complex word of each sentence is marked in red. We choose the top six substitution candidates for each algorithm. "Labels" is annotated by turkers. Words generated by the algorithms that are exist in the labels are highlighted in bold.}
\label{Example2}
\end{table*}

We also present the qualitative evaluation of simplification candidates. We randomly choose 10 short sentences from LexMTurk as examples. Table \ref{Example2} shows the top six candidates generated by the three approaches. We chooses the state-of-the-art two baselines based word embeddings (Glava{\v{s}}\protect\cite{glavavs2015simplifying} and Paetzold-NE \protect\cite{paetzold2017lexical}) as comparison. Candidates that exist in labels (annotated by people) are highlighted in bold. 

From Table \ref{Example2}, we observe that BERT-LS achieves the best simplification candidates for complex words compared with the two baselines based word embeddings. The baselines produce a large number of spurious candidates since they only consider the similarity between the complex and the candidate regardless of the context of the complex word. The candidates generated by BERT-LS can maintain the cohesion and coherence of a sentence without the need of morphological transformation.

\subsection{FULL LS Pipeline Evaluation}

In this section, we evaluate the performance of various complete LS systems. We adopt the following two well-known metrics used by these work \cite{horn2014learning,paetzold2017survey}. 

\begin{table*}
\centering
\begin{tabular}{l|cc|cc|cc}
\hline
& \multicolumn{2}{|c|}{LexMTurk}  & \multicolumn{2}{|c|}{BenchLS}  & \multicolumn{2}{|c}{NNSeval} \\
 &  Precision & Accuracy &  Precision & Accuracy & Precision & Accuracy   \\
\hline
Yamamoto & 0.066& 0.066  &  0.044 & 0.041 & 0.444 & 0.025 \\
Biran &0.714 & 0.034  &  0.124  & 0.123 & 0.121 & 0.121 \\
Devlin & 0.368 &  0.366 &  0.309  & 0.307 &  0.335 & 0.117  \\
Paetzold\-CA & 0.578 & 0.396  &  0.423 & 0.423 & 0.297 & 0.297    \\ 
Horn & 0.761 & 0.663  & 0.546  & 0.341 & 0.364 &  0.172 \\
Glava{\v{s}} & 0.710 & 0.682  & 0.480  & 0.252 & 0.456 & 0.197   \\
Paetzold\-NE & 0.676 &  0.676 & \textbf{0.642} & 0.434 & \textbf{0.544}  & 0.335  \\
\hline
BERT-LS & \textbf{0.776} & \textbf{0.776} & 0.607 & \textbf{0.607} & 0.423 & \textbf{0.423}\\
\hline
\end{tabular}
\caption{Full pipeline evaluation results using Precision and Accuracy on three datasets.}
\label{pipeline}
\end{table*}

\begin{figure*}
\begin{minipage}{0.3\linewidth}
\centerline{\includegraphics[width=6cm]{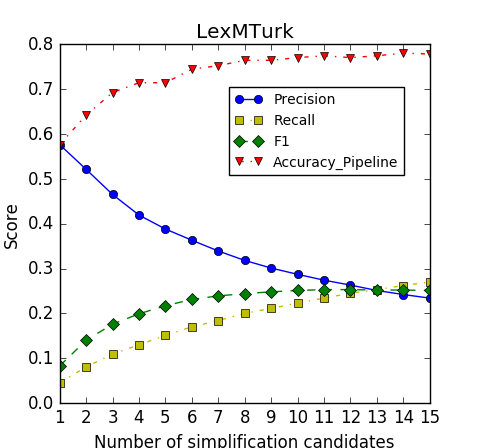}}
\end{minipage}
\hfill
\begin{minipage}{.3\linewidth}
\centerline{\includegraphics[width=6cm]{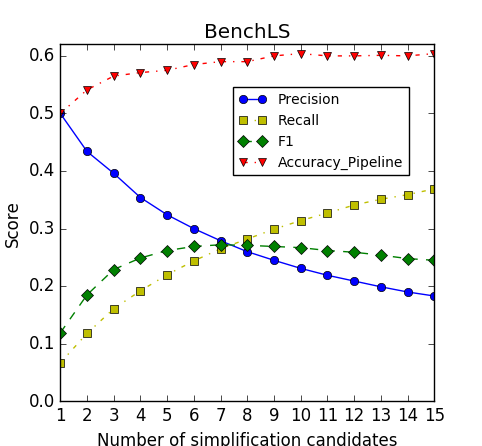}}
\end{minipage}
\hfill\begin{minipage}{0.3\linewidth}
\centerline{\includegraphics[width=6cm]{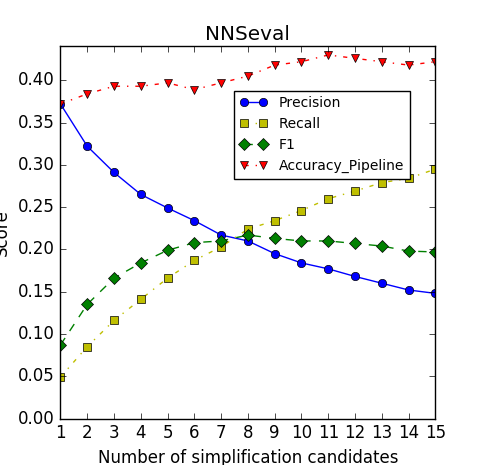}}
\end{minipage}
\caption{ Influence of number of simplification candidates.} \label{parameter}
\end{figure*}

\textbf{Precision}: The proportion with which the replacement of the original word is either the original word itself or is in the gold standard

\textbf{Accuracy}: The proportion with which the replacement of the original word is not the original word and is in the gold standard.

The results are shown in Table \ref{pipeline}. The results of the baselines on LexMTurk are from \cite{paetzold2017survey} and the results on BenchLS and NNSeval are from \cite{paetzold2017lexical}. We can see that our method BERT-LS attains the highest Accuracy on three datasets, which has an average increase of 12\% over the former state-of-the-art baseline (Paetzold-NE). It suggests that BERT-LS is the most proficient in promoting simplicity. Paetzold-NE is higher than BERT-LS on Precision on BenchLS and NNSeval, which means that many complex words in two datasets are replaced by the original word itself, due to the shortage of simplification rules in parallel corpora. In conclusion, although BERT-LS only uses raw text for pre-trained BERT without using any resources, BERT-LS remains the best lexical simplification method. 

\subsection{Influence of the Number of Simplification Candidates}

In this part, we try to investigate the influence of the number of simplification candidates to the performance of BERT-LS. The number of candidates ranges from 1 to 15, respectively. Figure \ref{parameter} shows the performance of simplification candidates (Precision, Recall and F1) and full LS pipeline (Accuracy) with different number of candidates on three benchmarks. When increasing the number of candidates, the score of precision decreases and the score of recall increases. When increasing the number of candidates, the score of F1 first increases, and declines finally. The best performance of simplification candidates through the experiments is achieved by setting the number of candidates equals 7 to 11 for a good trade-off between precision and recall. The score of the accuracy of full LS pipeline first increases and converges finally, which means that the whole LS method is less sensitive to the number of candidates. 

\section{Conclusion}

We proposed a simple BERT-based approach for lexical simplification by leveraging the idea of masking language model of BERT. Our method considers both the complex word and the context of the complex word when generating candidate substitutions without relying on the parallel corpus or linguistic databases. Experiment results have shown that our approach BERT-LS achieves the best performance on three well-known benchmarks. Since BERT can be trained on raw text, our method can be applied to many languages for lexical simplification. 
 
One limitation of our method is that it can only generate a single-word replacement for complex words, but we plan to extend it to support multi-word expressions. In the future, the pre-trained BERT model can be fine-tuned with just simple English corpus, and then we will use fine-tuned BERT for lexical simplification.

\section{Acknowledgment}
This research is partially supported by the National Key Research and Development Program of China under grant 2016YFB1000900; the National Natural Science Foundation of China under grants 61703362 and 91746209; the Program for Changjiang Scholars and Innovative Research Team in University (PCSIRT) of the Ministry of Education, China, under grant IRT17R32; and the Natural Science Foundation of Jiangsu Province of China under grant BK20170513.

\bibliographystyle{named}
\bibliography{AAAI-QiangJ.1590}

\end{document}